\documentclass[final]{cvpr}

\usepackage{times}
\usepackage{epsfig}
\usepackage{graphicx}
\usepackage{amsmath}
\usepackage{amssymb}
\usepackage{float}

\usepackage{cite}
\usepackage[utf8]{inputenc} %
\usepackage[T1]{fontenc}    %
\usepackage{url}            %
\usepackage{booktabs}       %
\usepackage{amsfonts}       %
\usepackage{nicefrac}       %
\usepackage{microtype}      %
\usepackage{bbm}
\usepackage{footnote}

\usepackage{pythonhighlight}

\usepackage{graphicx}
\usepackage{subcaption}
\usepackage{color, soul}

\usepackage[pagebackref=true,breaklinks=true,colorlinks,bookmarks=false]{hyperref}

\def\cvprPaperID{****} %
\def\confYear{CVPR 2021}

\newcommand{\xhdr}[1]{\vspace{6pt}{\noindent{\textbf{\textit{#1}}}}}

\begin{document}

\title{Reverse-engineer the Distributional Structure of Infant Egocentric Views for Training  Generalizable Image Classifiers}

\author{Satoshi Tsutsui    \hspace{5mm}   David Crandall\\
Indiana University\\
{\tt\small \{stsutsui,djcran\}@indiana.edu}
\and
Chen Yu\\
The University of Texas at Austin\\
{\tt\small chen.yu@austin.utexas.edu}
}

\maketitle

\begin{abstract}
   We analyze egocentric views of attended objects from infants. This
   paper shows 1) empirical evidence that children's egocentric views
   have more diverse distributions compared to adults' views, 2) we can
   computationally simulate the infants' distribution, and 3) the
   distribution is beneficial for training more generalized image
   classifiers not only for infant egocentric vision but for
   third-person computer vision.
\end{abstract}

\section{Introduction}

Children are highly efficient learners, and better understanding how
they succeed at visual learning could help build better machine learning
and computer vision systems.
Based on this ambitious motivation, we
have a long-running project to apply egocentric vision for
infants. The project has already provided many insights both for
developmental psychology~\cite{tsutsui2020wordlearning} and machine
learning~\cite{Bambach2018}. Other groups are also
investigating similarly-motivated
studies~\cite{orhan2020self,vong2021cross}, reflecting the
increasing interest in the intersection of egocentric vision and
infant learning.

\begin{figure}[t!]
    \centering
        \includegraphics[width=0.9\columnwidth]{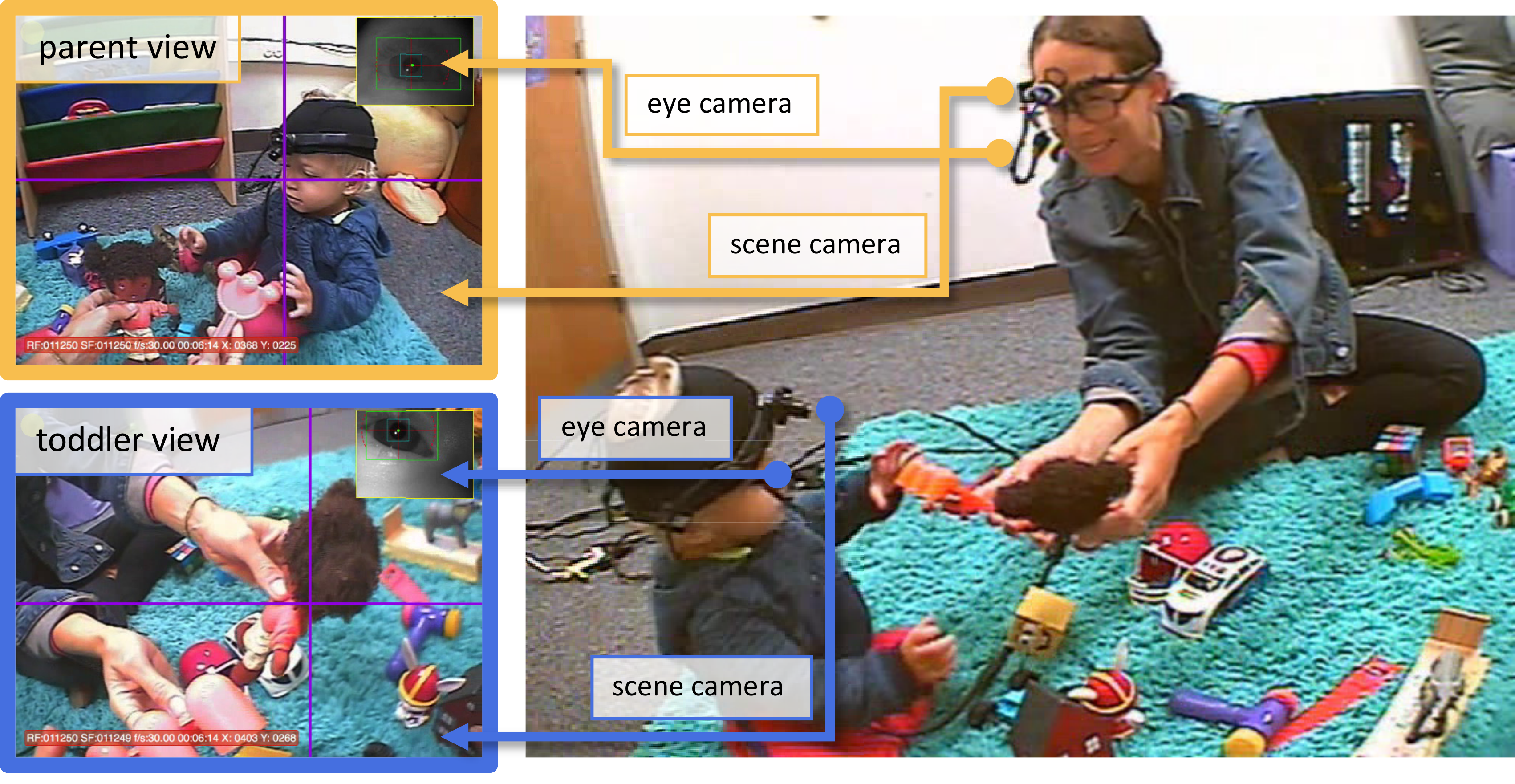}
    \caption{\textbf{Our experimental setup.} Child-parent dyads played together with a set of toys in a naturalistic 
    environment, while each wore head-mounted cameras to collect egocentric video and eye gaze positions (left).
    A stationary camera recorded from a third-person perspective (right).}
    \label{fig:setup_toys} 
    \vspace{-10mm}
\end{figure}

To study infant visual learning in everyday environments, 
we have collected 
egocentric video and eye gaze tracking data from
children and their parents as they freely play with 24 toy objects
(Figure~\ref{fig:setup_toys}). The wearable cameras provides an
approximation of the child's field of view --- the ``training data'' that they
use to learn object models.
We study the properties of this ``training data,'' for example using
it to train CNNs.
We find that deep networks trained from child views
perform significantly better than parent counterparts recorded in
exactly the same environment. We refer to our prior work~\cite{Bambach2018}
for more details.

This manuscript provides the latest findings from our project. 1)
Egocentric images collected from children have a unique distributional
property compared to adults. 2) We can computationally simulate image
classification datasets with the child-like distributional
property. 3) Image classification datasets with the child-like
property can train classifiers with higher generalization ability than
those without it.

\begin{figure}[t!]
	\centering
	\captionsetup[subfigure]{aboveskip=-33pt,belowskip=-10pt}
	\begin{subfigure}[b]{0.48\columnwidth}
		\includegraphics[width=\textwidth, clip]{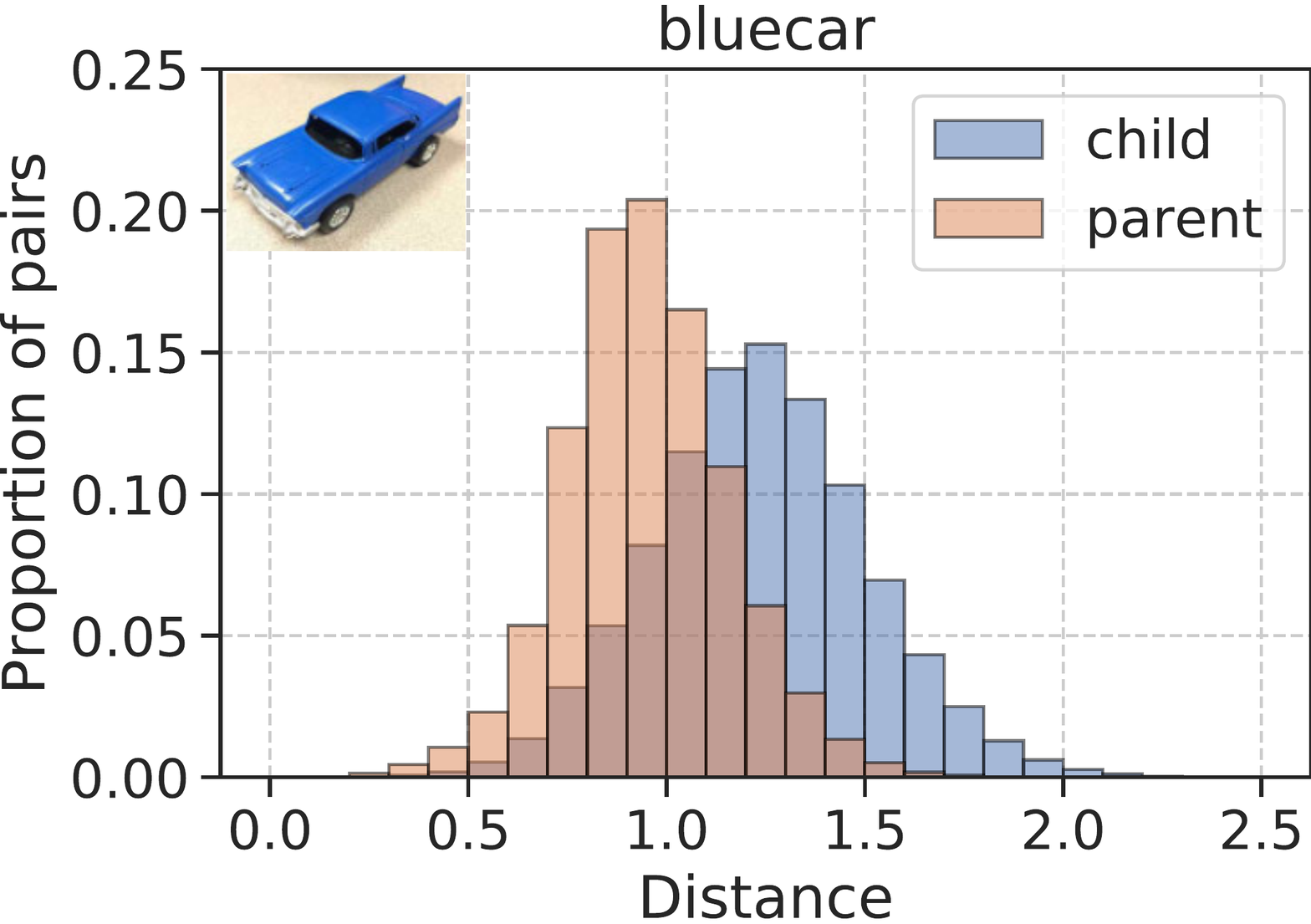}
		\caption{} 
		\label{fig:gist-pair-car}
	\end{subfigure}
	\hspace{4pt}
	\begin{subfigure}[b]{0.48\columnwidth}
		\includegraphics[width=\textwidth, clip]{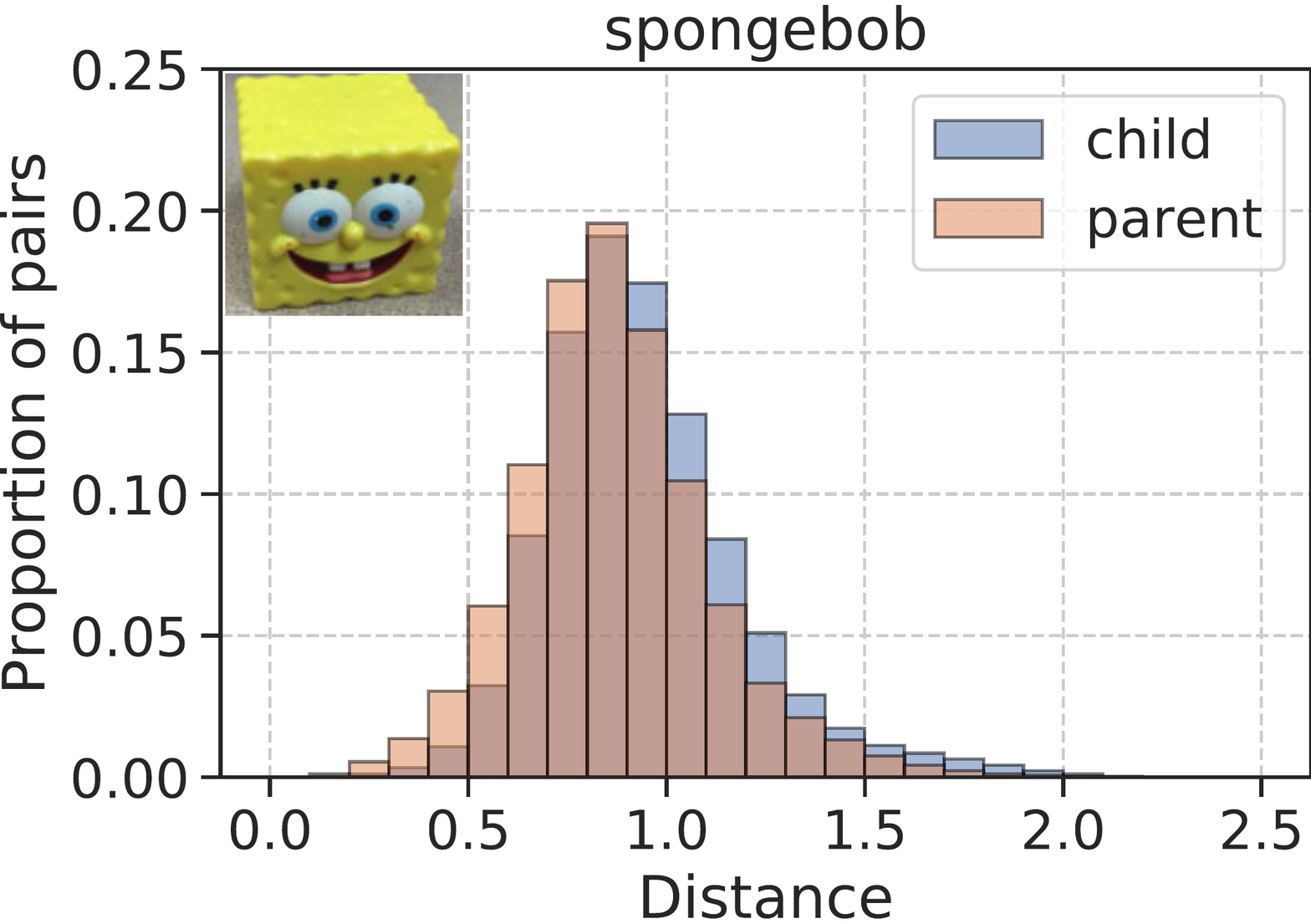}
		\caption{}
		\label{fig:gist-pair-bob}
	\end{subfigure}
	\caption{Histograms quantifying visual diversity of cropped instances for two specific objects: (a) the blue car and (b) the Sponge Bob toy.}
\end{figure}

\begin{figure}[t!]
	\centering
	\captionsetup[subfigure]{aboveskip=0pt,belowskip=-10pt}
	\begin{subfigure}[b]{0.48\columnwidth}
		\centering
		\includegraphics[width=\textwidth]{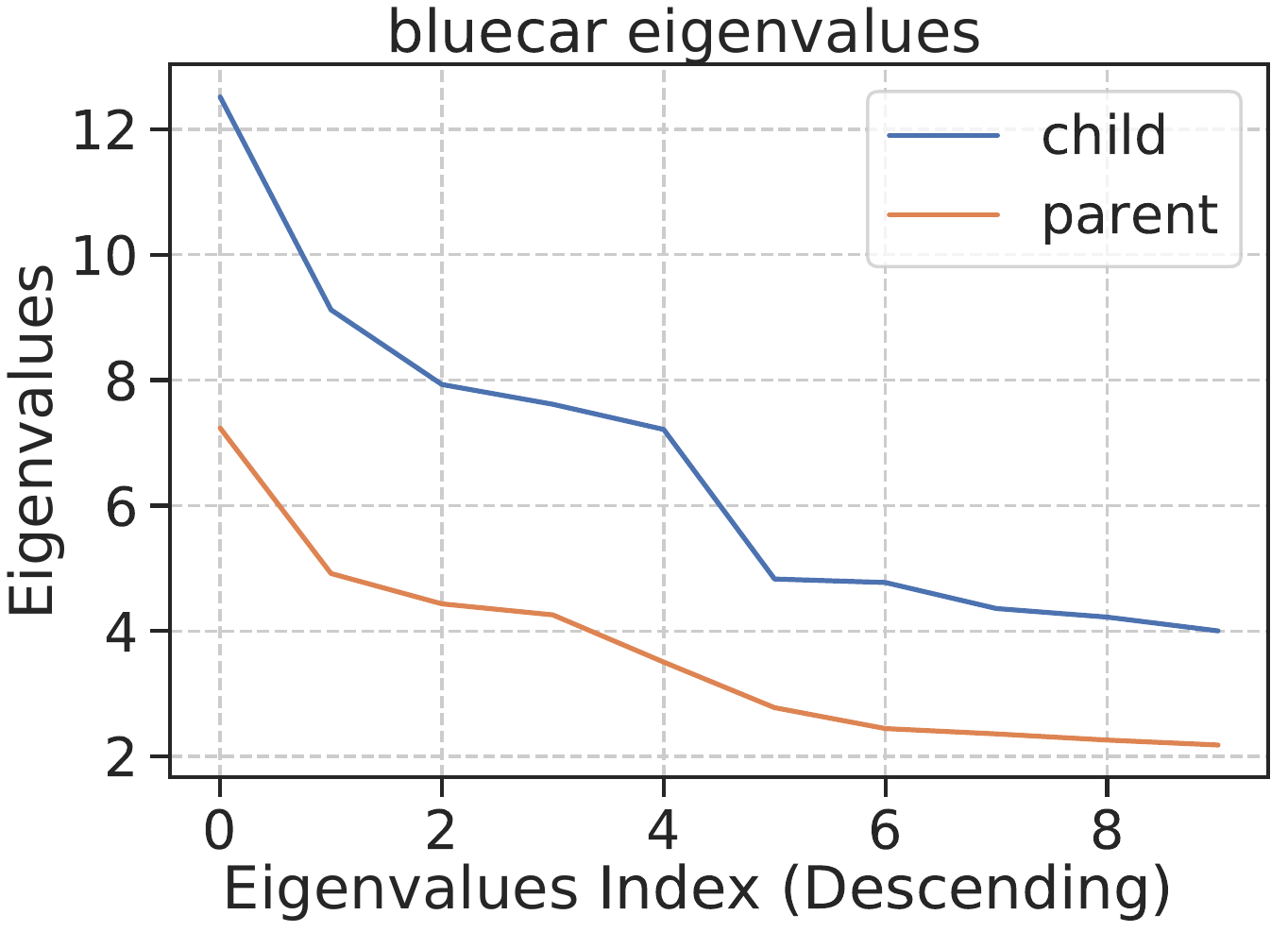}
		\caption{}
		\label{fig:pca-blue-car} 
	\end{subfigure}
	\begin{subfigure}[b]{0.48\columnwidth}
		\centering
		\includegraphics[width=\textwidth]{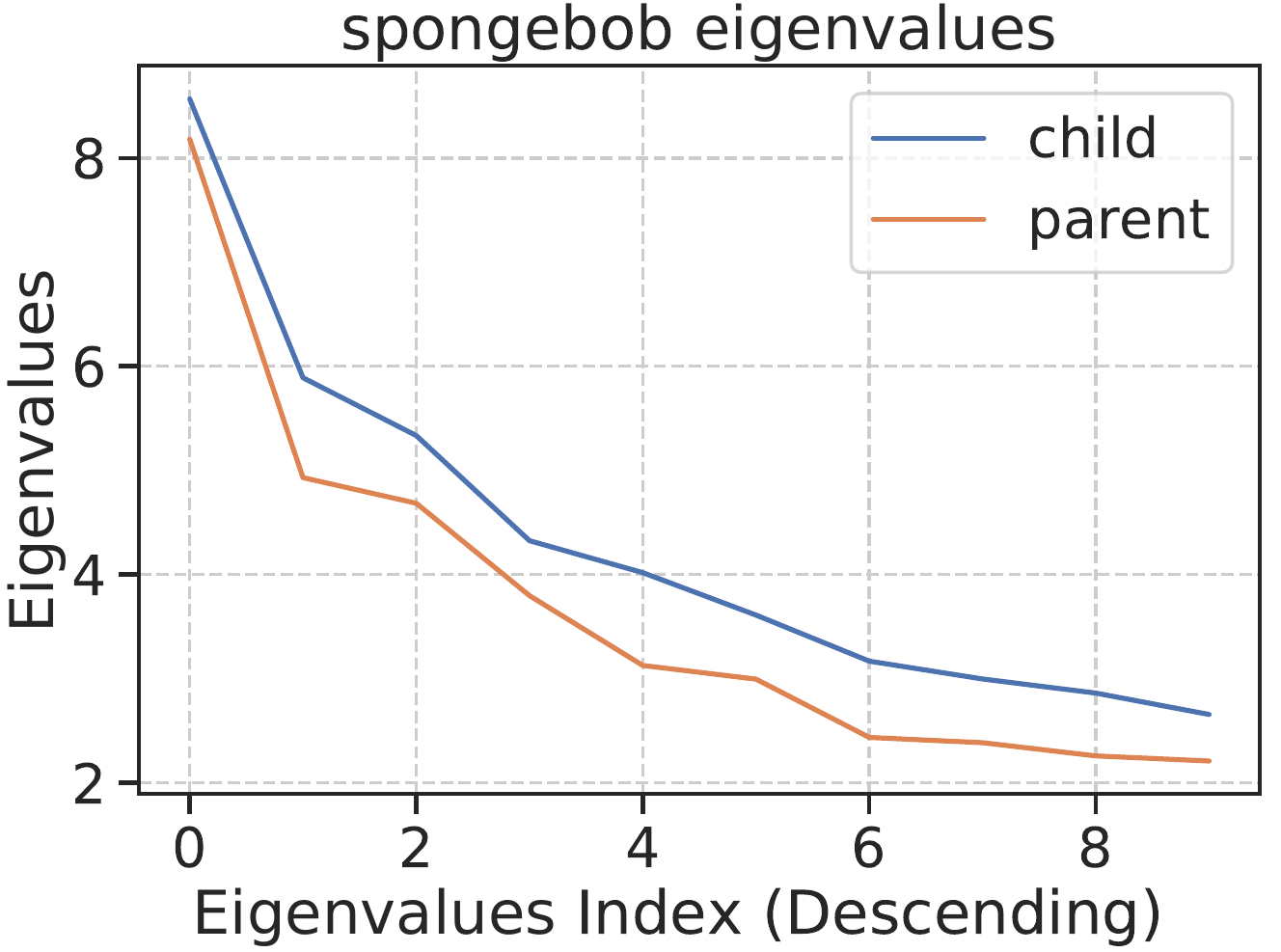}
		\caption{}
		\label{fig:pca-bob} 
	\end{subfigure}
	\caption{Another way of quantifying visual diversity is through eigenvalues of the principal components of PCA. Again, we see significantly greater variance for the child views for  (a) the blue car and (b) the Sponge Bob toy.}\label{fig:pca}
\end{figure}

\begin{figure*}[t!]
  \centering
          \begin{subfigure}[b]{0.32\textwidth}
    \centering
    \includegraphics[width=1.01\textwidth,trim=0cm 0.075cm 0 0cm,clip]{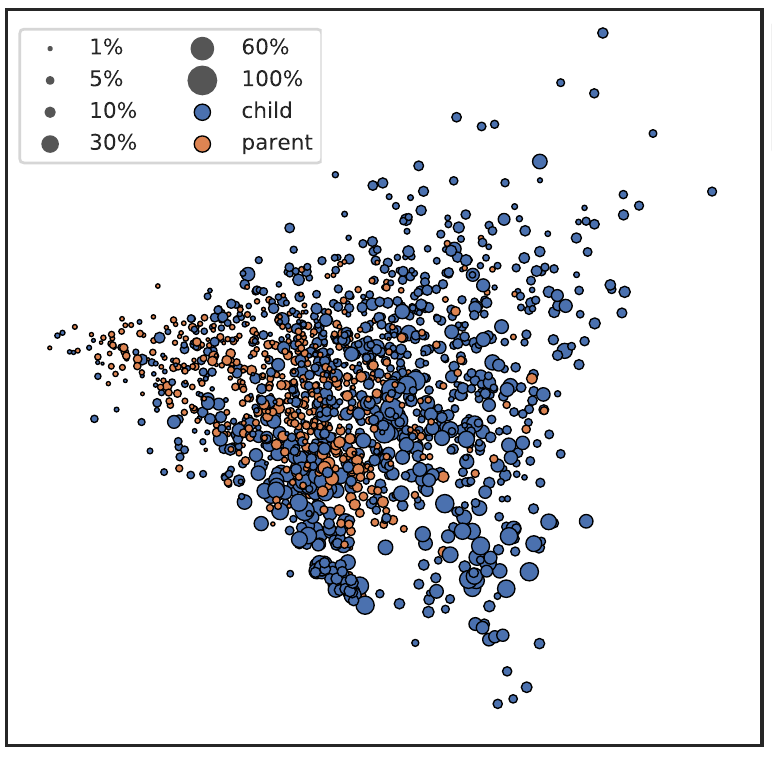}
    \caption{Combined}
    \label{fig:combineddots}
  \end{subfigure}
  \begin{subfigure}[b]{0.32\textwidth}
    \centering
    \includegraphics[width=\textwidth]{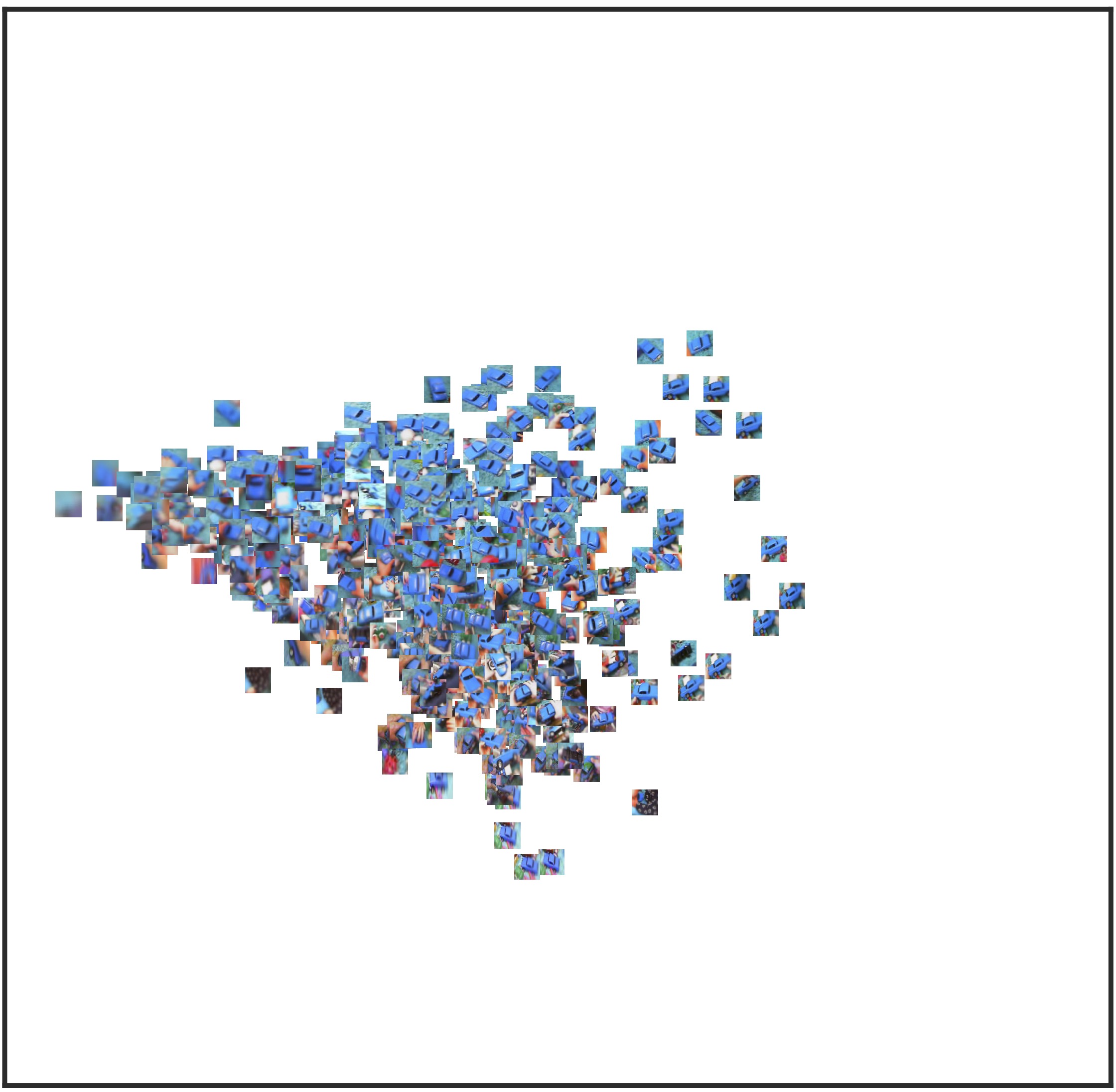}
    \caption{Parent views}
    \label{fig:parentmds}
  \end{subfigure}
    \begin{subfigure}[b]{0.32\textwidth}
    \centering
    \includegraphics[width=\textwidth,trim=0cm 0.1cm 0 0,clip]{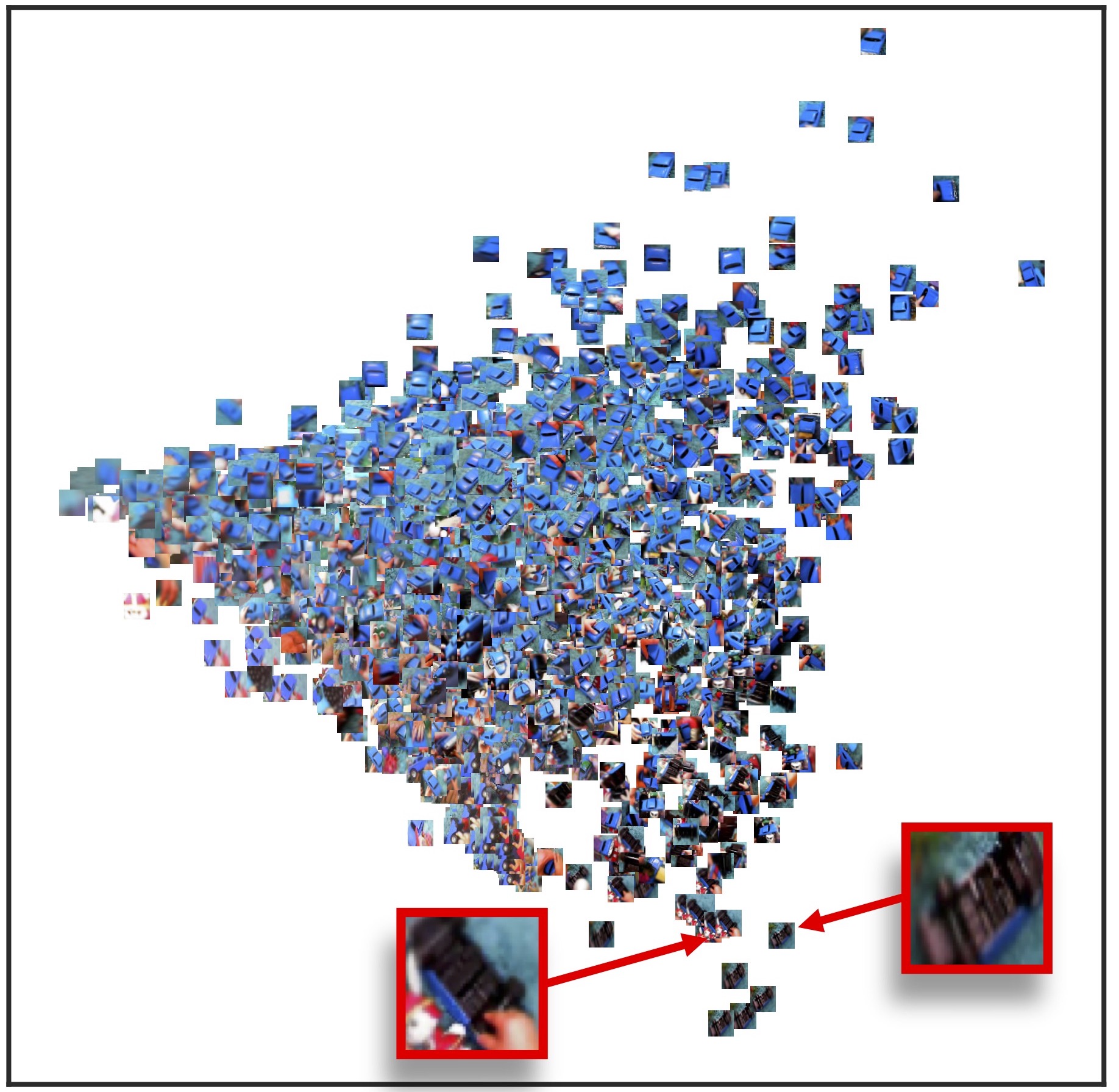}
    \caption{Child views}
    \label{fig:childmds}
    \end{subfigure}
    \caption{\textbf{Visualization of views of objects seen by parents
        and children} for one particular toy object, the blue car. The
      visualizations are 2D embeddings produced with Multidimensional
      Scaling (MDS) on GIST features of cropped images. (a): Each dot
      represents an image seen by parents (orange) or children (blue),
      and the size of the dot indicates the size of the object within
      the observer's field of view. (b) and (c): The same MDS plot
      with actual images superimposed on the dots, and split across
      (b) parents and (c) children. The children see
      considerably greater diversity in both size and visual
      appearance.}
	\label{fig:blue-car-mds-compare}
\end{figure*}

\section{Findings}
\subsection{Children's egocentric views indeed have a unique distributional property.}
Our previous work~\cite{Bambach2018,tsutsui2019active} suggests that views attended by children has more diverse distribution of object views with larger object size. In this section, we provide more empirical support for the previous findings. 

\xhdr{Distribution of pairwise distances} One way of quantifying the diversity of an image set is to compute the distance between all possible pairs of images, and investigate the distribution of the distances. We compute the Euclidean distances of all possible pairs of images for infants' attended views and parents' attended views for each toy. When computing a distance, we use GIST~\cite{gist} feature as image representation. We chose GIST because it is a low-level 
feature (as opposed to more semantic deep features) that is sensitive to the 
spatial orientation of an object; we wanted a distance metric such that 
two instances of the same object viewed 
from similar angles would have a small distance, while different 
views of the same object would have a large distance. Figures \ref{fig:gist-pair-car} and \ref{fig:gist-pair-bob} show the pairwise distribution plots of the two specific
toy objects: the blue car and the Sponge Bob doll.  We see that for
both objects, the children's data is more visually diverse than the
parents', but the difference is much more significant for the blue
car. This indicates that the diversity of
visual appearance is largely caused by greater diversity of viewing
angles, since the car has more complex 3-D structure
(e.g. body, tires, windows, etc) while the Sponge Bob is simpler and
looks very similar from different views. 

\xhdr{Variance from PCA} Another of quantifying visual diversity is through Principal Component Analysis (PCA),
which helps to interpret a high-dimensional space by 
decomposing it into orthogonal subspaces based on the variance of the data in the
original space.  We applied PCA for each object in the parent and child data and
examine the top 10 eigenvectors, which indicate the degree of visual diversity along
each of the principal components. We find that the eigenvalues of the child data are consistently higher than those of parent data; Figure \ref{fig:pca} shows two examples for  the blue car and the Sponge Bob doll. This means that the manifold of views attended by children is larger than the parent counterpart, which means children's data has more diversity. 

\xhdr{Visualizing visual diversity} To visualize the diversity of visual data created by adults and
children, we used Multidimensional Scaling (MDS) to embed the images
into a 2D space.   Figure~\ref{fig:combineddots} presents an MDS visualization
for the blue car, in which color indicates parent or child, and size of
the point indicates the size of the cropped object bounding box compared
to the viewer's field-of-view. Figures~\ref{fig:parentmds} and~\ref{fig:childmds}
present the same MDS plots, but with the actual images superimposed
on the points, and split across parent and child views. These plots confirm our previous findings.  First, child data
tends to have larger-sized objects, as the child data points (blue
dots in Figure~\ref{fig:combineddots}) tend to be larger than the
parent data points (orange dots). Second, child data has significantly
greater diversity than parent data, as the child data points in
Figure~\ref{fig:childmds} are more spread out than the parent data
points in Figure~\ref{fig:parentmds}.  For both children and parents,
there is a central ``core'' of highly-similar, typical, canonical
views of the object, but the children also see ``outlier'' views, such
as a cluster of images of the bottom of the car in the lower-right of
Figure~\ref{fig:childmds} that do not appear in the parent views.

\begin{figure}[t]
    \centering
    \includegraphics[width=\columnwidth]{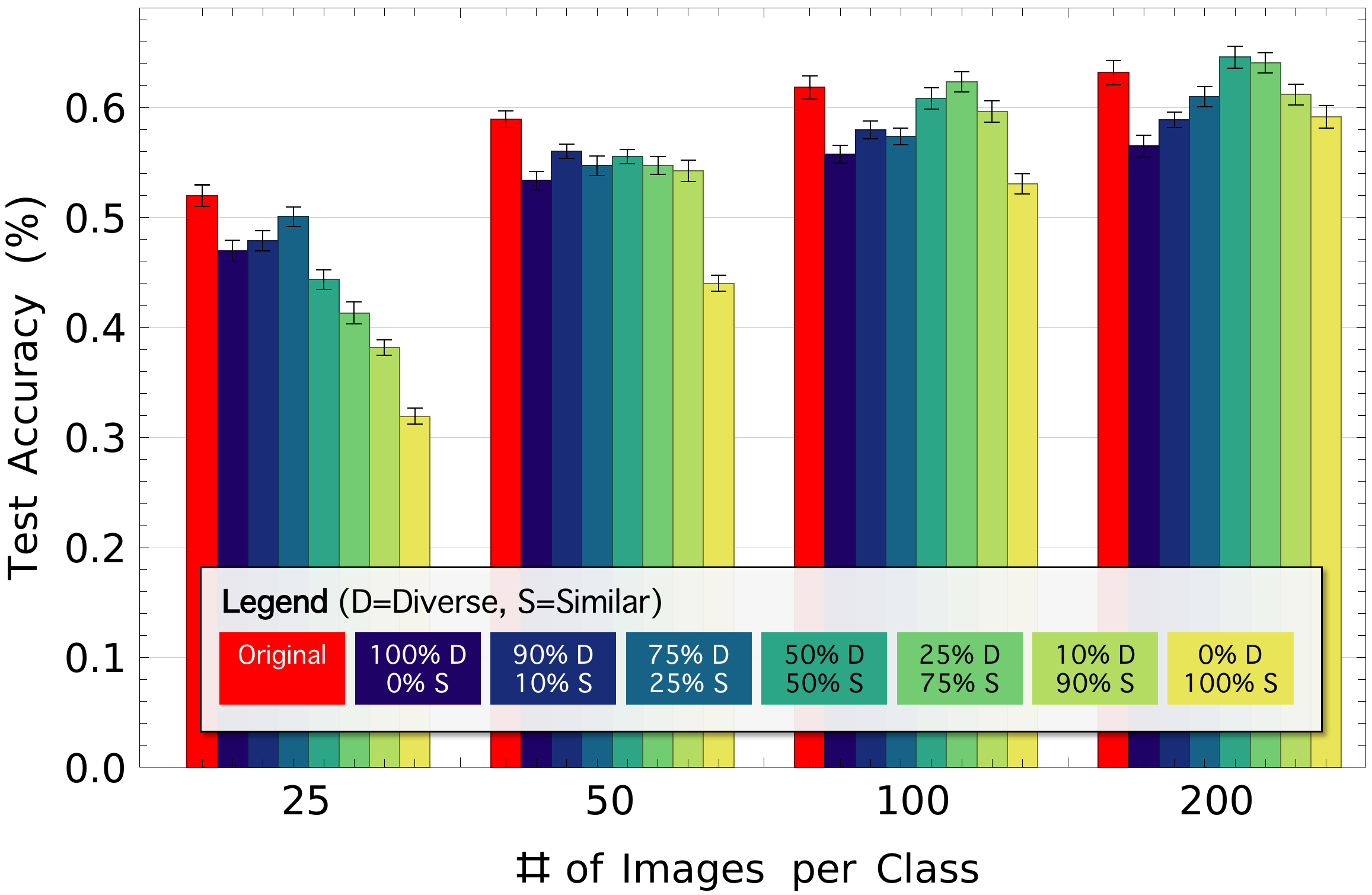}
	\caption{\textbf{Results of training on various mixtures of
        diverse and similar child egocentric data} while testing on
      independent third-person images, as a function of number
    of examples per class. Training on purely diverse (dark blue)
    or purely similar (yellow) subsets leads to significantly less
    accurate classifiers than the original child data (red), but a
    mixture of about 75\% similar and 25\% diverse leads to accuracy
    that is nearly as good. }\label{fig:child-mixbar}
\end{figure}

\subsection{Reverse-engineering the structure of child data}
Having confirmed that children's attended views have a unique diversity compared to the parent counterpart, we attempt to ``reverse-engineer'' the structure of the children's data so that we can apply the findings to provide better insights for data collection for training image classifiers. We proceed by trying to synthetically generate a training dataset that works
as well as the infant dataset, by artificially controlling the proportion of images that contribute to dataset diversity and those that do not. We approximate these sets as  diverse set and similar set using pairwise GIST distances. (The definitions of diverse/similar images are provided in Sec. 4.3 of our previous work~\cite{Bambach2018}).  Specifically, we created new datasets consisting of a fraction $p$ of randomly-sampled images
from the similar subset, and fraction $1-p$ of random images from the diverse subset.
We used gaze-centered crops 
with  $30^\circ$ field of view, and employed the same setup as previous work~\cite{Bambach2018} except that instead of producing a single set across all object categories,
we select subsets for each class and then produce class-balanced datasets.

We train CNNs (VGG16) using these subsets and compute accuracy on the held-out test images of the same objects in the third-person canonical view for 24-way object classification (See our previous work~\cite{Bambach2018} for details). Figure~\ref{fig:child-mixbar} presents accuracy for training datasets subsampled
to have different numbers of exemplars per class (25, 50, 100, and
200) with different proportions of diverse and similar images. We see that training sets consisting only of diverse images lead
to significantly better results than those consisting only of similar
training sets (e.g., about 52\% versus 32\% for 25 images per class),
until the number of images per class reaches 200. This is because when
there are 200 images per class, the similar set is itself quite diverse.

More importantly, we see that for any number of exemplars per class,
a mixture of diverse and similar sets always performs significantly better
than either set alone. 
This suggests that a high-quality training set needs
both similar and diverse training instances.
Moreover, for the dataset
size of 100 and 200 examples per class, the subsets consisting of 75\% similar images and 25\%
diverse images are as good as the original sets. This complements the
finding in previous work~\cite{Bambach2018} -- the data created by
toddlers, which consists of a mix of both similar and dissimilar
instances, is a unique combination of clustering and variability that
may be optimal for object recognition.  Indeed, we note that for the
dataset size of 25 and 50, the original set outperforms the any
combination of similar and diverse set. This suggests that the
combination of similar and diverse sets is not the only
characteristics that makes the child data better, and how toddlers
collect data efficiently in the data-scarce situation is interesting
future work.

\begin{figure*}[htb!]
	\centering
        \begin{tabular}{cc}
	\includegraphics[width=0.45\textwidth]{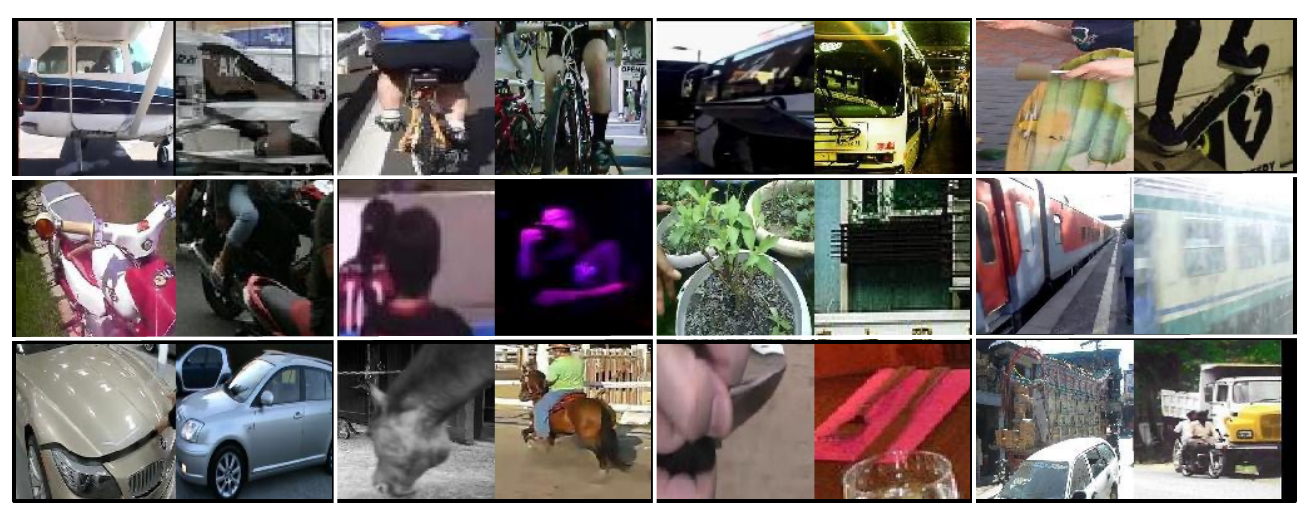} &
        	\includegraphics[width=0.5\textwidth]{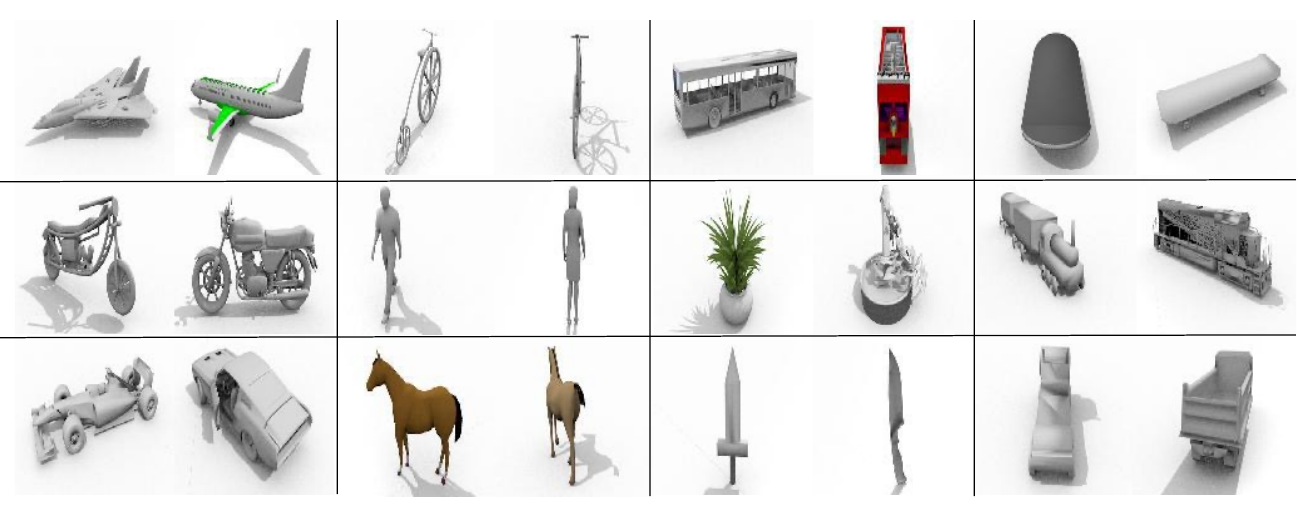}               \\
                (a) Sample training images & (b) Sample test images \\
                \end{tabular}
	\caption{\textbf{Sample images from the dataset for object model generalization experiments,} consisting of (a) training images from COCO and (b) test images from ShapeNet.
        Each cell in the figure shows two randomly-selected images from one of the 12 object categories.        }
	\label{fig:cvdataset}
\end{figure*}

\begin{figure}[t]
    \centering
    \includegraphics[width=\columnwidth]{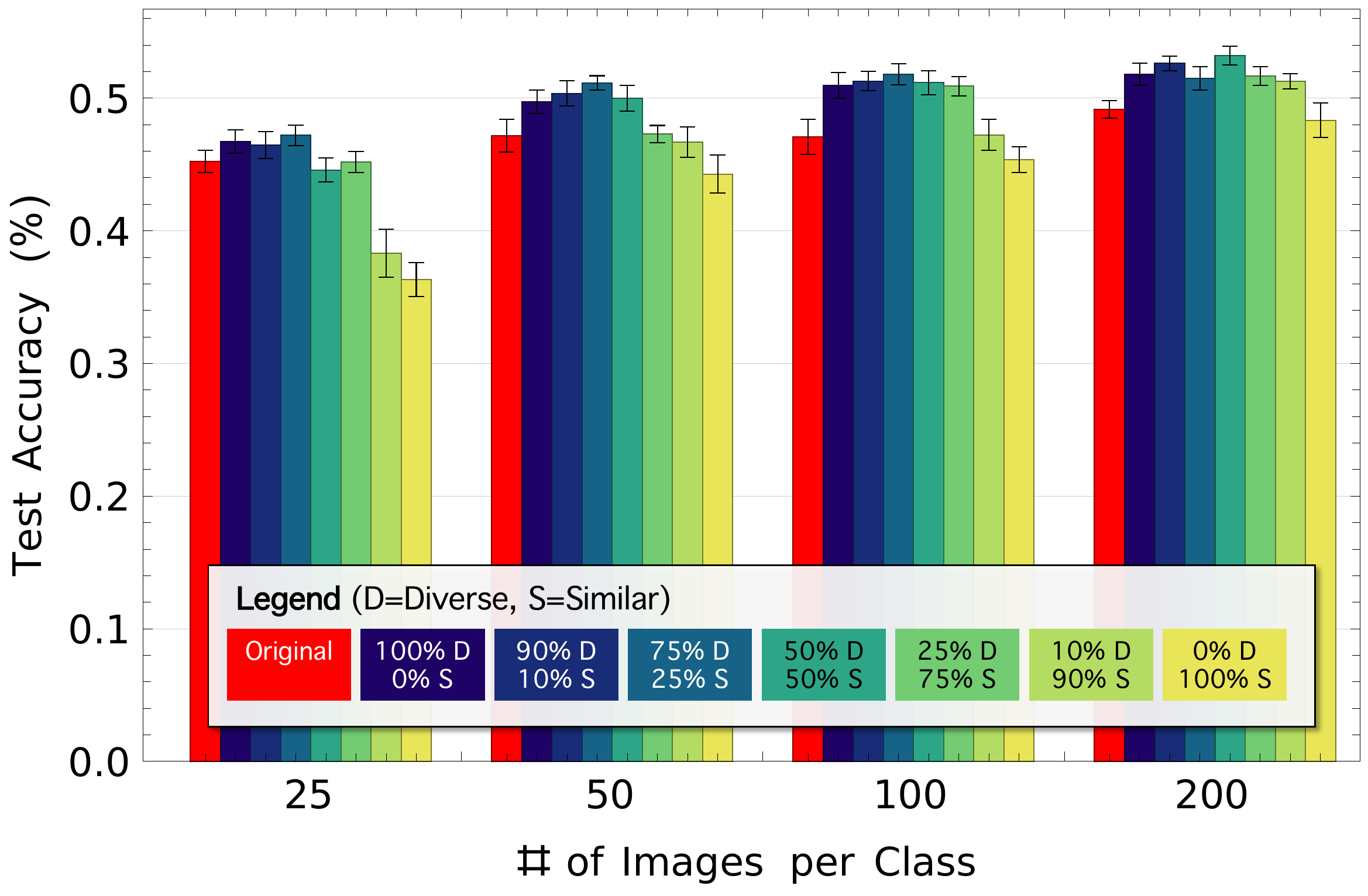}
	\caption{\textbf{Results of training on various mixtures of
        diverse and similar subsets of COCO} while testing on ShapeNet
        images.  A mixture of similar and diverse subsets leads to
        better accuracy on ShapeNet than the original COCO
        distribution, suggesting that a training distribution like that
        of the child data leads to more generalizable classifiers. }\label{fig:cv-mix}
\end{figure}

\subsection{Generalizing insights from child data}\label{sec:sim-div-general}

Inspired by the dataset from toddlers, the previous section shows a
key factor that makes the toddler data better -- combinations of
similar and diverse images. Can this same insight be used to collect
more generalizable training datasets in computer vision?

The vast majority of recognition datasets in computer vision include training and test
splits that are sampled from the same dataset. In contrast, we
need a dataset that can test our hypothesis that specific combinations
of diverse and similar images in training could lead to better generalization
in testing.  This, of course, is the way in which children are able to generalize
from, say, playing with toy firetrucks to recognizing real firetrucks as they drive by.

To do this, we constructed a dataset where the training data is from
natural images while the test set is from canonical images of the
objects. We collected training images from the MS COCO
dataset, and test images from ShapeNet corresponding to
the abstract representation of the objects.  The
dataset has 12 classes (aeroplane, bicycle, bus, car, horse, knife,
motorcycle, person, plant, skateboard, train, and truck). Each class
has 4,000 training images per class, totaling 48,000 images. We also
have 1,500 test images and 1,500 validation images per class, totaling
18,000 images for each set.  Figure~\ref{fig:cvdataset} shows some
sample images.

We note a key difference between this and the toyroom dataset task
above: that task considered object instance recognition (identical
objects for training and testing), but here we consider
the more challenging and realistic problem of category recognition.

We performed similar experiments on this dataset as
we did for child data, and show the results in
Figure~\ref{fig:cv-mix} as a function of number of images per class.
As with the child data, the results on this dataset show that training
datasets consisting only of diverse images lead to significantly
better accuracy than those consisting only of similar images. 
In addition, the
best accuracy is a combination of similar and diverse images, meaning
that we need both similar images, which possibly help create a
prototype representation, and diverse images, which help to capture
the representation of less typical cases.
A notable difference from the child results is the accuracy of random
subsets. Random subsets are inferior to the best combination of
similar and diverse images. This suggests that random sampling,
which is often used in computer vision work, is not always the best
strategy.

\section{Conclusion}
We provided empirical observations that egocentric object views attended by infants are more diversely distributed than parent views. Then, we try to simulate the child-like distribution in a computational way both for the original egocentric dataset and an image classification dataset. Our ultimate goal is to generalize the insights that we found from the study of toddler's visual development with egocentric vision into many computer vision problems, and we hope this paper shows the first step. 

{\small
\bibliographystyle{ieee_fullname}
\bibliography{references}
}

\end{document}